\documentclass[runningheads]{llncs}

\usepackage[year=2026,ID=2629]{eccv}

\usepackage{eccvabbrv}

\usepackage{graphicx}
\usepackage{booktabs}

\usepackage[accsupp]{axessibility}  %

\usepackage{bbm}

\usepackage{tcolorbox}
\tcbuselibrary{breakable}

\newcommand{\KA}[1]{{\color{black}#1}} %

\newcommand{\OURS}{\textsc{SportSkills}\xspace}
\usepackage{xurl}

\usepackage{booktabs}
\usepackage{array}
\newcommand{\PreserveBackslash}[1]{\let\temp=\\#1\let\\=\temp}
\newcolumntype{C}[1]{>{\PreserveBackslash\centering}p{#1}}
\newcolumntype{L}[1]{>{\PreserveBackslash\raggedright}p{#1}}

\newcommand{\hide}[1]{}

\usepackage[pagebackref,breaklinks,colorlinks,citecolor=eccvblue]{hyperref}

\usepackage{orcidlink}

\begin{document}

\title{\OURS: Physical Skill Learning from \\ Sports Instructional Videos} 

\titlerunning{\OURS: Physical Skill Learning from Instructional Sports Videos}

\author{Kumar Ashutosh, Chi Hsuan Wu, Kristen Grauman}

\authorrunning{Ashutosh et al.}

\institute{University of Texas at Austin}

\maketitle

\begin{abstract}

Current large-scale video datasets focus on general human activity, %
but lack depth of coverage on fine-grained activities needed to   %
address physical skill learning. %
We introduce \OURS, the first large-scale sports dataset geared towards physical skill learning with in-the-wild video. \OURS has more than $360k$ instructional videos containing more than $630k$ visual demonstrations paired with instructional narrations explaining the know-how behind the actions from 55 varied sports. 
Through a suite of experiments, we show that \OURS unlocks the ability to understand fine-grained differences between physical actions. Our representation achieves gains of up to $4\times$ with the same model trained on traditional activity-centric datasets. 
Crucially, building on \OURS, we introduce %
the first large-scale task formulation of \emph{mistake-conditioned instructional video retrieval}, bridging representation learning and actionable feedback generation (e.g., ``here's my execution of a skill; which video clip should I watch to improve it'').
Formal evaluations by %
professional coaches show our retrieval approach significantly advances the ability of video models to personalize visual instructions for a user query. Project: \href{https://vision.cs.utexas.edu/projects/sportskills/}{\url{https://vision.cs.utexas.edu/projects/sportskills/}} 
  \end{abstract}

\section{Introduction}
\label{sec:intro}

Understanding human actions has long been a key focus area in computer vision. The field has progressed from human action classification~\cite{omnivore,mvitv2,uniformer,memvit,slowfast},  anticipation~\cite{avt,rulstm,intention,whenwillyoudowhat,gao2017red}, pose estimation~\cite{wham,vitpose,sun2019deep,newell2016stacked,toshev2014deeppose}, human motion tracking~\cite{girdhar2018detect,doering2018joint,bertasius2019learning,li2021tokenpose}, human/hand-object interaction~\cite{gkioxari2018detecting,tamura2021qpic,wang2020learning,kim2021hotr,ning2023hoiclip}, to action quality assessment~\cite{egoexo4d,action-quality-assessment,baller-gedas,stl-vs-mtl,interpretable-feedback,basket-gedas}, and recently, human action feedback~\cite{expertaf,qevd,crosstrainer,exact-dataset-gedas}. The underlying technical challenge in all these tasks is to understand the difference between human actions. While the difference in actions is drastic for distinct actions, \eg, dancing vs sleeping, the distinction is minimal but crucial as we move towards physical skill assessment.

\begin{figure}[t]\footnotesize
    \centering
    \includegraphics[width=\linewidth]{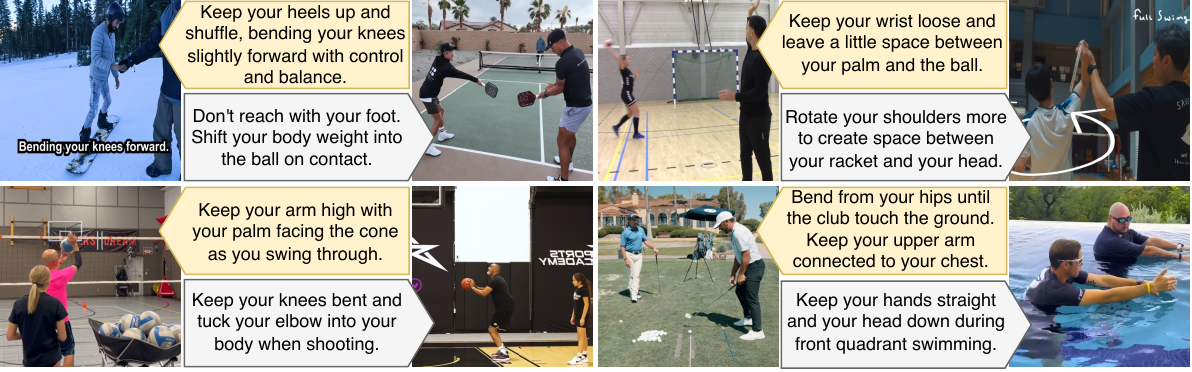}
    \includegraphics[width=\linewidth]{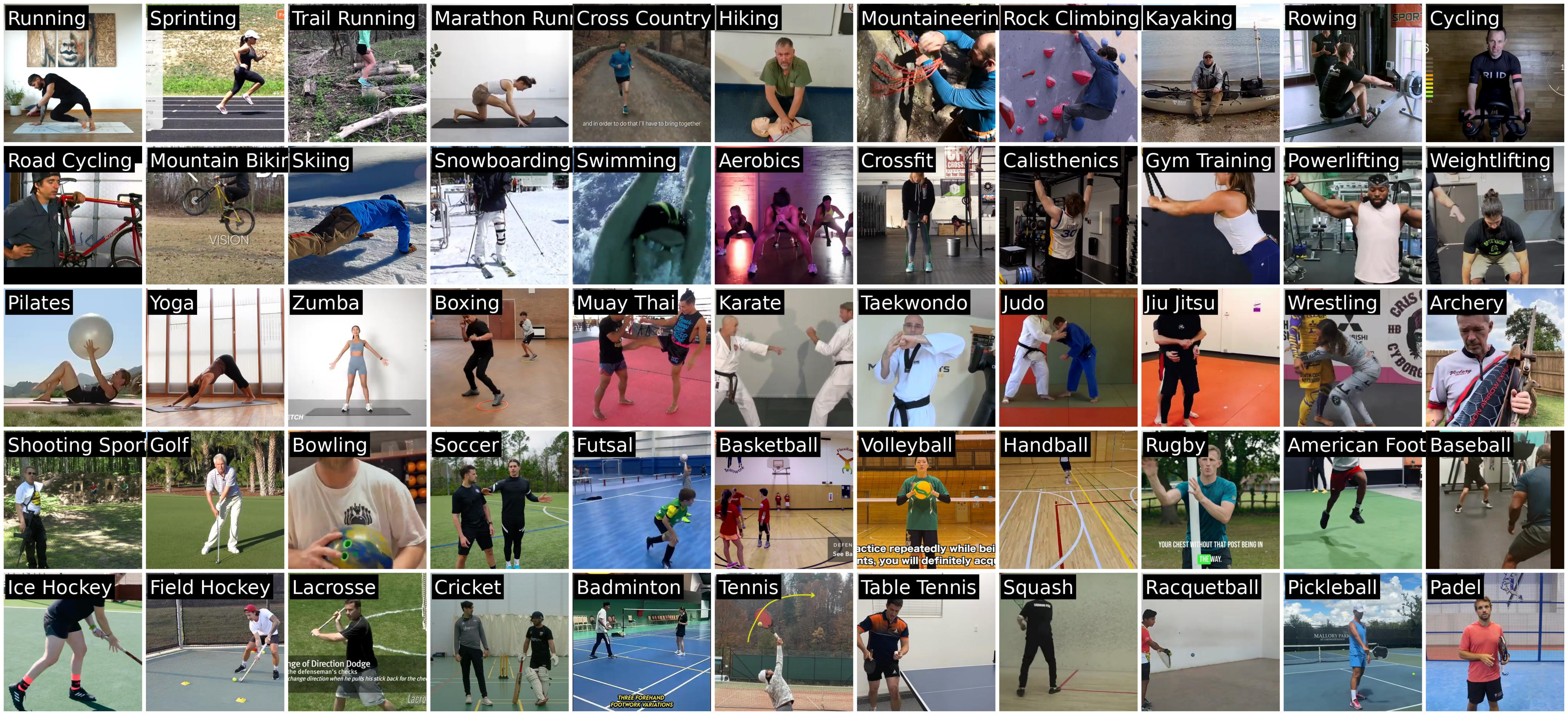}
    
    \caption{Our proposed \OURS, sourced from YouTube~\cite{youtube}, contains paired videos and instructional narrations describing the correct execution of a skill (top), totaling $638,399$ video clips. All the visuals are meant for skill-improvement. We propose this dataset to enable video models to learn physical skills. \OURS contains $55$ popular sports (bottom), ranging from judo, sprinting, to tennis and handball.}
    \label{fig:teaser}
    \vspace{-0.5cm}
\end{figure}

The key challenge in perfecting physical skill understanding is to learn the fine-grained difference between executions. For example, the difference between a correct and an incorrect soccer dribble is subtle, and often only spotted by an experienced player or a coach. Enabling AI to attain similar performance requires visual expertise that must be fed to the models. In short, current video models %
lack physical skill understanding.
A key reason for this limitation is that the 
current datasets~\cite{howto100m,kinetics-600,ego4d,ucf101} used in video models %
lack depth of coverage of such physical actions, %
particularly sports. 
Consequently, today's video representations are general-purpose, but lack the ability to find subtle physical action differences, as also demonstrated in our experiments.

In this work, we solve the technical challenges in obtaining a stronger model for physical skill understanding. We address the problem on all fronts: we propose a large-scale sports dataset for physical skill learning, a video representation trained on the collected dataset, and a key application putting the dataset to use in %
AI coaching for skill improvement.

\KA{Firstly, to address the shortage of data in physical skill learning, we introduce \OURS, a large-scale collection of sports instructional videos. We focus on this domain because sports expertise is fundamentally defined by subtle, fine-grained physical differences, and instructional videos explicitly capture these distinctions through visual demonstrations of correct techniques.}
The dataset, sourced from YouTube~\cite{youtube}, contains video demonstrations that are paired with expert instructions about specific drills. It has $638,399$ paired video clips, obtained from a pool of $369,296$ videos spanning $55$ popular physical sports. We curate the dataset from the raw videos %
using state of the art large language models (LLM)~\cite{llama3modelcard} and vision language models (VLM)~\cite{qwen25vl} to isolate clips describing a concept, along with a visual demonstration, \eg, dribbling a ball in soccer. Our proposed dataset is $13\times$ larger than the closest dataset~\cite{basket-dataset-gedas} in duration and offers order(s) of magnitude better volume and depth compared to existing resources for physical skills (see \cref{tab:dataset_compare}). %
See \cref{fig:teaser} for samples. \KA{Crucially, unlike prior datasets, our dataset is explicitly constructed to capture fine-grained variations in skill execution that define expert physical performance.}

Our proposed dataset unlocks learning better physical skill video representations. Specifically, we demonstrate the inability of existing video encoders in associating fine-grained skill narrations with the visual demonstrations.
\KA{In contrast, our trained video representation excels and can classify better into correct and incorrect skill execution with minimal linear probing.  Training with \OURS improves physical skill understanding by as much as $4\times$.}

\KA{Finally, we demonstrate a central application enabled by \OURS: personalized skill coaching through instructional video retrieval. Effective skill learning requires more than retrieving a generic instructional video—the content must address a learner’s specific mistakes, skill level, and style so that the feedback is directly actionable. However, existing query-based video retrieval systems are not designed for this setting and typically return videos based on high-level semantic similarity~\cite{internvideo2,videoclip,howto100m} rather than the fine-grained technique differences relevant for skill improvement. In contrast, we formulate the novel task of \emph{mistake-conditioned instructional retrieval}: given a learner’s video that contains errors or suboptimal execution, our proposed method retrieves instructional clips that best demonstrate the correct execution of the skill in a way that directly addresses the learner's mistakes. %
This capability is uniquely enabled by \OURS, which pairs fine-grained demonstrations with expert instruction. To evaluate this setting, we conduct a study with professional sports coaches who annotate the test set. Our method significantly surpasses the strongest existing baseline by $10\%$.}

In summary, this work advances physical skill understanding by introducing \OURS and demonstrating its effectiveness through a suite of challenging tasks, and we open up a real pathway for personalized skill learning from in-the-wild how-to videos.  %
The dataset and baseline codes will be released to the community to support continued research in this rapidly growing space.

\section{Related Work}
\label{sec:related_work}

\noindent \textbf{Instructional and sports video datasets.} Instructional video datasets~\cite{howto100m,crosstask,coin,youcook2} are of great interest to the research community for their usefulness in learning video representations~\cite{howto100m,mil-nce,hero,videoclip}, task execution~\cite{detours,stitch-a-recipe,procedure-learning-fei-fei-li,procedure2,procedure3}, and procedural knowledge~\cite{paprika,task_graph,video-distant}.
They mostly focus on cooking, DIY, and gardening. While these datasets are critical in learning complicated multi-step procedures, the focus is not physical, \ie, exactly how the activity is executed is not central. 
\KA{In particular, HowTo100M~\cite{howto100m} is an inspirational resource containing 1M videos (100M clips), and has accelerated research in procedural understanding and enabled many  tasks~\cite{detours,procedure-learning-fei-fei-li,htstep-neurips2023,stepdiff}. However, the dataset does not focus on physical skills; the largest dataset for physical skills is $30\times$ smaller~\cite{basket-dataset-gedas}.}

Sports video datasets~\cite{ahmidi2017dataset,skill-in-videos,interpretable-feedback,parmar2017learning,mtlaqa,finediving,zhang2023logo,skating-eval,baller-gedas,ego-exo-learn,basket-dataset-gedas,exact-dataset-gedas, SPORTS1M} (see~\cref{tab:dataset_compare} for comparison) are 
used for action classification, temporal localization, or skill assessment. Recently, Ego-Exo4D~\cite{egoexo4d} and QEVD~\cite{qevd} go beyond skill-level classification, and \KA{introduce videos paired with expert commentary. However, these datasets are manually captured, and hence, the scale is $102\times$ smaller than an internet-scale dataset~\cite{howto100m}.}
\KA{Taking inspiration from the success of HowTo100M~\cite{howto100m} for procedural tasks, we propose \OURS, sourced from YouTube~\cite{youtube}, as the first dataset with paired sports demonstrations and expert narrations. 
The strong visual demonstrations enable learning skill-aware video models.}

\noindent \textbf{Sports assessment benchmarks.} Prior work focusing on sports understanding predicts current action~\cite{soomro2015action,ucf101,thumos,wu2022survey,KarpathyCVPR14}, action phase~\cite{skill-in-videos,soccernet,finegym}, or an execution score/skill-level~\cite{egoexo4d,action-quality-assessment,baller-gedas,stl-vs-mtl,interpretable-feedback,basket-gedas,quality-distribution-learning,group-aware-regression}, a mistake~\cite{finediving,zhang2023logo,whos-better-dima}, or actionable feedback~\cite{expertaf,qevd,crosstrainer}. 
Out of these, predicting the action's quality score or  the person's skill-level is good for assessment but not useful for skill improvement. That is, telling a learner they did $8/10$ will not help them find areas of improvement. To handle this issue, recent work proposes mapping a learner's video to actionable feedback~\cite{expertaf,qevd,egoexo4d}. %
While language feedback is strictly more helpful in skill learning than a score, %
it is well-known that humans often learn better visually~\cite{buch2014video}.
\KA{ExpertAF~\cite{expertaf} does propose demonstration retrieval as a task, but the videos are not instructional in nature, and hence, not truly feedback videos.}
Hence, we propose a method and a benchmark enabling a visual-feedback task: mapping a learner's video to the most relevant instructional video clip.
Going beyond prior work, 
we propose a method to retrieve instructional clips as a form of feedback given a learner video. In summary, this method will further strengthen the use of AI in sports coaching and skill understanding.

\begin{figure}[t]\footnotesize
    \centering
    
    \includegraphics[width=\linewidth]{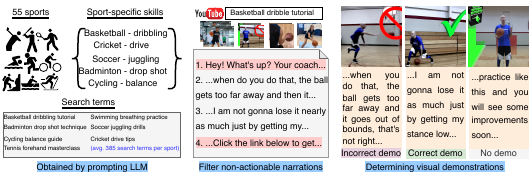}
    
    \caption{\textbf{\OURS collection overview.} 
   We develop a curation pipeline to create a solid skill-learning video-language dataset sourced from in-the-wild coaching/how-to videos.
    We prompt an LLM~\cite{gpt4} to generate a list of physical sports, and sport-specific skills. We use connecting words like `tutorial', `drills' to create search terms (left). Next, we query YouTube~\cite{youtube} with these queries and obtain narrations. We use an LLM~\cite{gpt4} to filter out non-actionable instances, \eg, ``click the link to get discounts...'' (middle). Finally, we use a VLM~\cite{qwen25vl} to obtain pairs of $(v, t)$ with correct or incorrect demonstrations, filtering out cases without visual demonstrations (right). 
    }
    \vspace{-0.5cm}
    \label{fig:data}
\end{figure}

\section{\OURS Dataset}
\label{sec:dataset}

We curate a physical instructional dataset $\mathcal{D} = \{(v_1, t_1, c_1), (v_2, t_2, c_2), ...\}$ such that $v$ demonstrates a sports skill, along with a spoken narration $t$ of the know-how of the skill, \eg a video with the correct golf swing posture, with the narration describing \emph{``bending the hips until the club touches the ground''}, see~\cref{fig:teaser} (top two rows). The demonstration can either show how to do it correctly $(c=1)$, or can show mistakes and what to avoid $(c=-1)$. Our goal is to obtain a large-scale clean set $\mathcal{D}$, such that $v$ and $t$ are strongly aligned, and $c$ is the right demonstration type.

We obtain videos from YouTube~\cite{youtube}, as frequently used as a video source in prior video datasets~\cite{howto100m,youcook2,pre-computed-1,crosstask,coin}. We consider a list of sports and the key concepts of that sport. We then perform a text-only filtering to obtain narrations that are instructional in nature. Finally, we use the video and the narrations with a vision language model (VLM) to label the pairs as showing an instructional skill. %
Our careful filtering ensures strongly aligned $(v, t)$ training pairs. See~\cref{fig:data}.
In the following, we first detail the data collection pipeline (\cref{sec:dataset-pipeline}), followed by our method of extracting visual actionable feedback instances (\cref{sec:dataset-filtering}). 

When choosing the sports to include in \OURS, we impose the following requirements: 1) It should be a popular sport, 2) We focus on sports requiring physical actionable feedback, unlike chess or non-visual concepts like \emph{mentality}, 3) Within a sport, we specifically focus on skills that are single-person, \eg, dribbling and shooting over collective team strategies. Single-person skills provide a cleaner action–outcome relationship, enabling more precise modeling of how fine-grained motion patterns translate into measurable performance improvements, as also explored in~\cite{egoexo4d,qevd}.

\subsection{Data collection pipeline}
\label{sec:dataset-pipeline}

\noindent \textbf{Obtaining list of sports.} Based on the above requirements, we query GPT-4o~\cite{gpt4}, a large language model (LLM), with the above requirements and obtain a list of 50 sports. The exact LLM prompt, and the resulting list of sports is provided in the Appendix. All the resulting sports are manually checked by the authors for the three conditions above, and we add $5$ more sports that we find missing in the obtained list.

\noindent \textbf{Obtaining focus skill-set.} We observe that learners need different skill-sets in different sports. Therefore, we 
curate the skills for each sport separately based on the above requirements, \ie, the skill should be visual and single-person focused. We again query the LLM~\cite{gpt4} to generate a list of skills central to learning the sport, along with our condition.

\noindent \textbf{Obtaining videos from YouTube.} We concatenate the sport (\eg soccer) and skill (\eg dribbling) with instructional keywords (\eg tutorial) to construct a variety of query terms (\eg soccer dribbling tutorial). We use the YouTube API to obtain top-100 results for each query, consistent with~\cite{howto100m}. Finally, we download the video and the English subtitles (consisting of the coach's spoken instructions) for extracting visual actionable feedback clips, as described next.

\subsection{Extracting \emph{visual} actionable feedback instances}
\label{sec:dataset-filtering}

\begin{table*}[t]
    \centering
        \caption{\textbf{Comparing \OURS with other datasets in skill learning.} Our dataset is $\mathbf{13\times}$ longer and $\mathbf{12\times}$ more videos than second longest dataset~\cite{basket-dataset-gedas}. 
        The number of skills and domains we consider are also significantly larger than prior datasets.}
    \resizebox{\textwidth}{!}{
    \setlength{\tabcolsep}{6pt} 

    \begin{tabular}{L{2.6cm}
C{2cm} C{3cm}
C{2cm} C{2cm}}
    \hline
    Dataset & Total Hours & \#Videos (\#Participants) & \#Skills & \#Domains \\ \hline
    JIGSAWS \cite{ahmidi2017dataset} & 3.5 & 103 (8) & 3 & 1 \\
    BEST \cite{skill-in-videos} & 26 & 500 (400) & 5 & 5 \\
    MIT-Dive \cite{interpretable-feedback} & 0.1 & 159 & 1 & 1 \\ 
    MIT-Skating \cite{interpretable-feedback} & 7.3 & 150 & 1 & 1 \\ 
    UNLV-Dive \cite{parmar2017learning} & 0.4 & 370 & 1 & 1 \\ 
    MTL-AQA \cite{mtlaqa} & 1.5 & 1,412 & 1 & 1 \\ 
    FineDiving \cite{finediving} & 3.5 & 3,000 & 1 & 1 \\ 
    LOGO \cite{zhang2023logo} & 11 & 200 & 1 & 1 \\ 
    Fis-V \cite{skating-eval} & 23.6 & 500 & 1 & 1 \\ 
    FP-Basket \cite{baller-gedas} & 10.3 & 48 & 1 & 1 \\
    EgoExoLearn \cite{ego-exo-learn} & 120 & 747 & 8 & 8 \\
    Ego-Exo4D \cite{egoexo4d} & 1,286 & 5,035 (740) & 8 & 8 \\
    BASKET \cite{basket-dataset-gedas} & 4,477 & 28,106 (32,232) & 20 & 1 \\
    ExAct \cite{exact-dataset-gedas} & 13.5 & 3,521 (155) & 8 & 8 \\ 
    QEVD-300K \cite{qevd} & 460 & 44,589 (1,900) & 148 & 1 \\ 
    QEVD-Coach \cite{qevd} & 4.5 & 223 (29) & 23 & 1 \\ 
    \hline
    \textbf{\OURS} & \textbf{57,314} & \textbf{349,141} (\textbf{220,761}) & \textbf{1,702} & \textbf{55} \\
    \hline
    \end{tabular}
    }
    \label{tab:dataset_compare}
    \vspace{-0.5cm}
\end{table*}

In addition to imparting knowledge, videos on YouTube are also meant to be enjoyable, welcoming, and easy to learn. Hence, these videos often contain color commentary like \emph{``welcome to my channel!''} and \emph{``this is my favorite drill''}~\cite{vnd,tan}. While these instances are helpful in keeping the viewer engaged, it's not helpful for video models. Secondly, many videos are low-quality, non-visual, or unrelated to coaching. Thus, we need to filter these instances to create visually-aligned video segments in \OURS. We next describe the pipeline to obtain high-quality visual clips that are helpful for skill learning.

\noindent \textbf{Finding actionable instructional instances.} Given a video and its subtitle, our objective is to obtain video segments, along with narrations, that are strictly visual and aimed at teaching concept(s). As discussed above, the undesired clips either do not contain any instructional content, or do not show them visually.

Firstly, we perform a text-only filtering to eliminate narrations that are non-visual. This process is intentionally text-only for efficiency, \ie, narrations like \emph{``welcome to my channel!''} are non-visual irrespective of the visual content. We use an LLM~\cite{gpt4} to classify narrations 
as \emph{actionable} or \emph{non-actionable}.

Next, note that even though a narration may be actionable, the actual visual content may not be aligned, \ie, it can show a coach's headshot speaking about a skill, or an unrelated or misaligned visual. We use a video-based vision language model (VLM), Qwen-2.5-VL-32B~\cite{qwen25vl}, to determine whether a given narration is visually depicted or not. Furthermore, even when it is visually demonstrated, the coach may either show how to do it correctly, or show common incorrect ways of doing it. The VLM is instructed to choose between these three categories: correct or incorrect demonstration, or no visual demonstration at all.

\noindent \textbf{Quality of the weakly-supervised dataset.} Note that the dataset is weakly-supervised, \ie, we do not use any manual supervision, and all the steps are guided by an LLM or a VLM. Nevertheless, our pipeline ensures minimal annotation error. Upon manual examination of $200$ narrations, we observe that $95\%$ of them are correctly labeled by the LLM~\cite{gpt4} as actionable and non-actionable. Next, out of $200$ outputs from the VLM~\cite{qwen25vl}, $91\%$ of them agree with the authors' assessment of whether the visual demonstration correctly, incorrectly, or does not show the narration. Moreover, our large-scale collection of $349,141$ videos, with $100$ videos per search query, helps mitigate biases in the skills being taught, teaching styles, geographical locations, and camera setups.

\subsection{Learning a physical skill-aware video representation}
\label{sec:pretraining}

After obtaining the \OURS dataset, we showcase improved skill-aware video representations enabled by it. Given $\mathcal{D} = \{(v_i, t_i, c_i=1)\}$ as a collection of clips where $v_i$ is a \emph{correct} demonstration of a skill, 
and $t_i$ is the corresponding narration text by the expert, we aim to learn a video representation function that correctly captures the fine-grained actions in $v$ and its distinction from all other actions $v'$ and narrations $t' \neq t$.

\textbf{Learning objective.} Following the success of contrastive learning in training video representations~\cite{videoclip,hiervl,egovlpv2,egovlp,howto100m,mil-nce,hero,clip}, we take a contrastive learning approach to learn the association between a sports video demonstration and its fine-grained narration. 
Given $(v_i, t_i)$ demonstrating a skill, we first use a state of the art encoders $\phi_v$ and $\phi_t$ \cite{clip,internvideo2} 
for encoding vision and text modalities. We add small trainable projectors $\pi_v$ and $\pi_t$ for video and narration, respectively, to enable fine-grained action understanding. In summary, we obtain normalized video and text embeddings as $\mathbf{v} = \pi_v(\phi_v(v))$ and $\mathbf{t} = \pi_t(\phi_t(t))$, respectively. Finally, we use the following learning objective to train the contrastive model

\begin{align*}
\mathcal{L} = -\frac{1}{2N}\sum_{i=1}^{N} \left[ \log \frac{\exp(\mathbf{v}_i^\top \mathbf{t}_i / \tau)}{\sum_{j=1}^{N} \exp(\mathbf{v}_i^\top \mathbf{t}_j / \tau)} + \log \frac{\exp(\mathbf{t}_i^\top \mathbf{v}_i / \tau)}{\sum_{j=1}^{N} \exp(\mathbf{t}_i^\top \mathbf{v}_j / \tau)} \right]
\end{align*}

\noindent where 
$N$ is the batch size, and $\tau$ is the temperature hyperparameter. The two terms are video to text and text to video contrastive losses~\cite{mil-nce,clip}, respectively. We show the pretraining and downstream task performance in the experiments.

\section{Providing Visual Instructional Feedback}
\label{sec:egoexo-to-youtube}

\KA{In this section, we introduce our proposed task of finding the right instructional video to learn and improve. Specifically, given a learner video with mistakes or scope for improvement, we design a model to provide relevant clips from \OURS that correctly convey improvements---a \emph{visual} correction, as opposed to language feedback. As motivated in the introduction, learning the right physical skills requires practice of the pose, execution, and motion. This task is also unique to physical skills and sports because the improvements and feedback have essential visual components, as opposed to procedural tips that are less visual and more `keyword searchable'.} 
For example, cooking feedback such as \emph{``cook longer until the sauce thickens''} or gardening advice like \emph{``water the plant less frequently''} can be easily expressed and retrieved through keywords, whereas correcting a tennis grip requires visual demonstration.

Specifically, given a learner video $d$ with suboptimal execution, we want to learn a scoring function $S$ to rate the usefulness of any instructional video $v$ for improving the suboptimal execution. That is, $S(v, d)$ will be high for $v$ that can help improve the learner's suboptimal executions
 in $d$. The overall video recommendation is $v^* = \arg\max_{v \in \mathcal{D}} S(v, d)$ for a retrieval set $\mathcal{D}$.  Since $\mathcal{D}$ contains paired $(v, t)$, we can convert this video-to-video ($d\rightarrow v$) problem to a computationally easier video-to-text ($d\rightarrow t$) problem, say $S'$. The corresponding problem is of finding $t^*$, such that $(v^*, t^*) \in \mathcal{D}$, and $t^* = \arg\max_{t \in \mathcal{D}}S'(t, d)$.

\subsection{Creating the train and test data}
\label{sec:train-test-for-visual-feedback}

There is no dataset that can be directly used for this task.
Thus, we curate a train and a test data, as described next.
\OURS provides us with pairs of $(v, t, c)$. However, we need pairs of $(d, v)$ for training and testing $S'$ 
(and hence $S$).   We obtain $d$ from Ego-Exo4D~\cite{egoexo4d}, which provides expert-annotated pairs of learner video and expert commentary, \ie $(d, a)$ where $a$ is the expert commentary. 
We use expert commentary only for dataset curation; our model does not observe any expert commentary during training or inference.
To match the right $v$ with the right $d$, we use weak supervision to create the train set, and we curate a gold-standard test set for testing. We obtain the test set from professional coaches, and it benchmarks future progress in this task.

\noindent \textbf{Training data.} From Ego-Exo4D~\cite{egoexo4d}, we have $a$ in text-form that are experts' feedback on sports skills. Also, we have $t$ in \OURS that are the YouTube narrations (in text-form) describing the visual demonstrations. Even though $a$ and $t$ are not the same thing, they do share semantic similarities. An instructional video narration asking to \emph{``bend the knees to have a better balance''} is likely a good visual recommendation for an expert's feedback to \emph{``bend the knees more''}. Thus, for a given $(d, a)$, we assign $(v^*, t^*)$ as the corresponding instructional video when $t^* = \arg\max_{t \in \mathcal{D}} \psi(t, a)$ where $\psi$ is a sentence similarity function~\cite{mpnet}.

\noindent \textbf{Testing data.} While the above idea of matching $a$ with $t$ helps obtain a weakly supervised training set, we cannot rely on this for testing. Automated similarity metrics are prone to error, and even narrations with high similarity scores are not guaranteed to \emph{coach} the same topic. 
Thus, we introduce a gold-standard testing dataset CoachGT for this task.  
We ask experienced professional sports coaches, each with $10+$ years of experience, 
to see the learner video $d$, read the actionable feedback $a$ provided by their fellow experts (in~\cite{egoexo4d}), and rate whether a candidate $(v, t)$ is very relevant, somewhat relevant, or not relevant. The experts are briefed about the task, and paid for their time. More details of the annotation process are given in the Appendix and ~\cref{fig:bench}. Overall, we create $451$ expert annotated pairs of $(v, d)$, along with the relevance annotation.
This manually annotated data will benchmark future developments in this task.

\subsection{Learning a feedback relevance function}
\label{sec:vlm_reranker}

\begin{figure}[t]\footnotesize
    \centering
    
    \includegraphics[width=\linewidth]{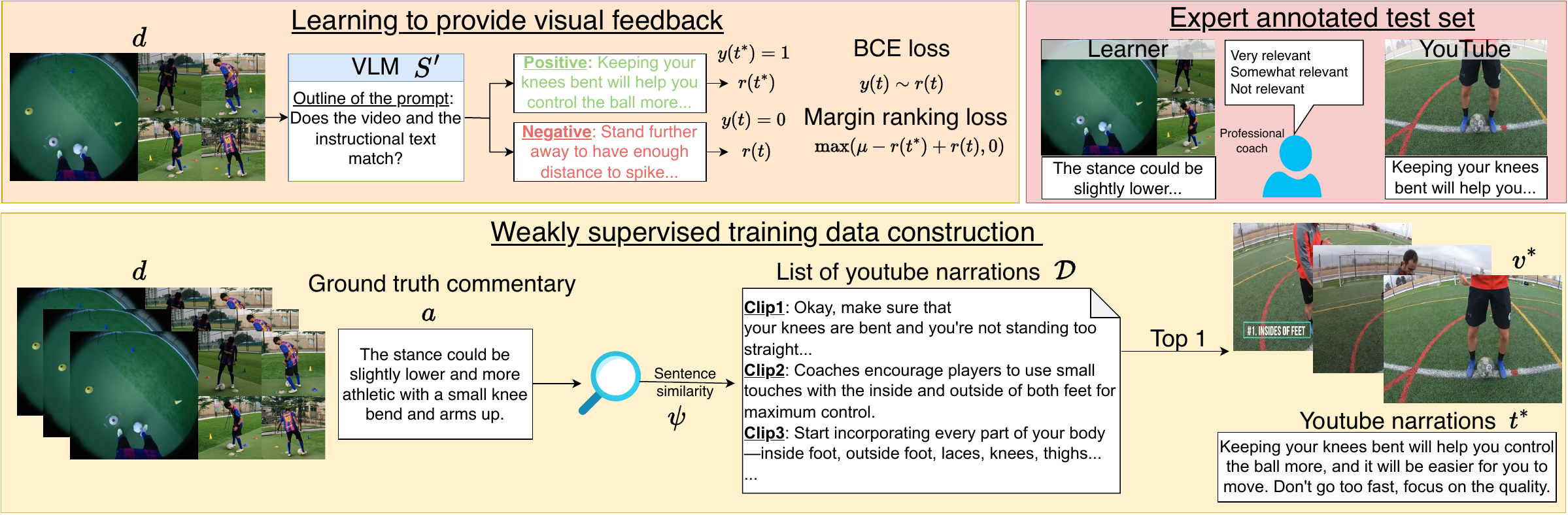}
    
    \caption{\textbf{Overview of our method} to retrieve visual feedback given a suboptimal execution by a learner (top left). We finetune a VLM~\cite{qwen25vl}, in LoRA~\cite{lora} setting, to predict if a given feedback candidate is relevant or not. 
    Top right: overview of the expert annotation task where they see the learner video, actionable feedback in text form, and the candidate YouTube video as visual feedback, and they rate the relevance of the pair to create CoachGT.  Bottom: our weakly supervised training data construction. 
    }
    \label{fig:bench}
    \vspace{-0.5cm}
\end{figure}

Recall that our objective is to learn a scoring function $S$, s.t. $S(v, d)$ is high when $v$ is the visual demonstration to correct the suboptimal execution $d$. Since \OURS has paired $(v, t) \in \mathcal{D}$, we learn a scoring function $S'(t, d)$ instead of our desired $S(v, d)$. 
We use a VLM~\cite{qwen25vl} for $S'$. The inputs to the VLM are the learner video $d$, and a candidate narration from a video-narration pair. The VLM outputs either `1' or `0' to
indicate the relevance of $t$.
We sample positives and negatives to train $S'$. The exact prompt for $S'$ is given in the Appendix.

\noindent \textbf{Learning objective.} We use a two-fold objective. Firstly, a simple binary cross entropy (BCE) loss $\mathcal{L}_{BCE}$ is designed to ensure $S'(t, d) = 1$ when $t=t^*$.
Secondly, we also use a margin ranking loss~\cite{marginranking-1,marginranking-2,marginranking-3} to ensure $S'$ for the positives are more than $S'$ for the negatives by at least  a margin $\mu$.

Formally, for a given instructional feedback candidate $t$, we define a relevance function $r(t) = \mathbb{P}\{S'(d, t) = 1\} - \mathbb{P}\{S'(d, t) = 0\}$. Since $S'$ is a VLM, $r(t)$ is simply the difference between probabilities of next token being `1' or `0'. Recall that `1' and `0' represent $t$ being the correct instructional feedback or not. 
Thus, the ranking loss can be written as
\begin{align*}
    \mathcal{L}_{rank} = \frac{1}{|\mathcal{N}|} \sum_{t \in \mathcal{N}} \max\left( \mu - r(t^*) + r(t), 0 \right),
\end{align*}
where $t^*$ is the positive sample, and $t\in \mathcal{N}$ is sampled from the negative set.  The overall training objective is $\mathcal{L} = \mathcal{L}_{BCE} + \lambda \mathcal{L}_{rank}$ for a hyperparameter $\lambda$. We discuss the implementation details in~\cref{sec:impl}.

\section{Experiments and Results}
\label{sec:expt_and_result}

We first discuss implementation details (\cref{sec:impl}), followed by our experiments in learning physical skill-aware video representations (\cref{sec:representation-learning}). Finally, we show results in providing visual feedback for a given learner video with suboptimal execution  (\cref{sec:visual-feedback-training}).

\subsection{Implementation details}
\label{sec:impl}

\noindent \textbf{Dataset statistics.} For our $55$ sports of interest, we collect a total of $1,702$ focus skills, \ie, an average of $31$ skills per sport. After querying YouTube for every search term, we prepare $546,479$ videos to download. Out of these videos, several of them do not have any narration, or are not available in English (even with auto-translate), or are not available for download. This results in $369,296$ downloaded subtitles. After filtering for LLM-only actionability using \cite{gpt4}, we have $341,075$ videos with at least one actionable narration. The resulting $638,399$ clips contain $559,962$ correct visual demonstrations and $78,437$ incorrect visual demonstrations. Throughout, we use GPT-4o~\cite{gpt4} for text-only processing, and Qwen-2.5-VL-32B~\cite{qwen25vl} as VLM.

\noindent \textbf{Video representation learning.} The trainable mappers $\phi_t$ and $\phi_v$ are simple neural networks with a single hidden layer. The hidden and the output layer are $256$-dimensional. We use three choices of visual encoders $\phi_v$~\cite{timesformer,clip,internvideo2}, compared in the result section, along with CLIP text encoder~\cite{clip} $\phi_t$. The temperature parameter $\tau$ is initialized with $0.07$ and it is learnable. We train a model separately for every sport. We train for $50$ epochs with a learning rate of $3\times 10^{-5}$ and a batch size of $64$ with AdamW~\cite{adamw} optimizer and a cosine learning rate scheduler. We use a single GH200 GPU for training and sample $8$ video frames per sample.

\noindent \textbf{Visual instructional feedback.} We use Qwen-2.5-VL-2B~\cite{qwen25vl} as the base model for $S'$. Furthermore, we apply LoRA~\cite{lora} fine-tuning to $S'$ with LoRA rank as $16$, $\alpha=32$ and $0.05$ dropout. The ranking margin $\mu$ is $0.2$, and the loss weighing constant $\lambda$ is $0.5$. The batch size is $4$, and we train the model for $10$ epochs. 
We sample hard-negatives for $\mathcal{N}$ by taking $t$ that have high similarity score with the learner video $d$, but $t \neq t^*$, \ie, it is not the ground truth for $d$.
The hardware and the sampled video frames are the same as above. In this task, we consider basketball, soccer, and rock climbing as we use Ego-Exo4D~\cite{egoexo4d} for obtaining learner videos $d$. Since the task is aimed at skill improvement, we only use feedback that are actionable, and ignore narrations $a$ that rate the action as perfect. Note that \cite{egoexo4d} does not provide such an annotation, and hence, we use Qwen-2.5-VL-32B~\cite{qwen25vl} to filter the narrations. We use 
a time-synchronized collage of all views (ego- and exocentric), provided in \cite{egoexo4d},
as input. 
The test set, CoachGT, contains $451$ samples, annotated by $6$ professionals ($2$ in each sport). See Appendix for details about the annotation process.

\subsection{Physical skill-aware representation learning}
\label{sec:representation-learning}

\begin{table}[t]
\centering
\footnotesize
\setlength{\tabcolsep}{3pt}

\caption{(Left) \textbf{Linear probe PR-AUC} of classifying a demonstration as correct or incorrect 
for soccer, basketball, and rock climbing. We compare CLIP ViT-B/32 with the same model trained with the \OURS dataset. Our learned representation gives better results.
(Right) \textbf{Evaluating visual instructional feedback.} We report average Cohen’s $d$ and pairwise concordance for the same sports. We see that our method outperforms both trained (Video Cont., CrossTrainer~\cite{crosstrainer}) and zero-shot baselines (ZS VLM~\cite{qwen25vl}). 
}

\begin{minipage}[t]{0.34\linewidth}
\centering
\begin{tabular}{
L{1.6cm}
C{1.0cm} C{1.0cm}
}
\toprule
& \multicolumn{2}{c}{Lin. Probe PR.} \\
\cmidrule(lr){2-3}
\textbf{Sport} & ViT-B/32 & w/ Ours \\
\midrule
Soccer   & 58.2 & \textbf{63.2} \\
Basketball  & 60.2 & \textbf{64.9} \\
Rock Cl. & 56.5 & \textbf{61.2} \\
\bottomrule
\end{tabular}
\end{minipage}\hfill
\begin{minipage}[t]{0.65\linewidth}
\centering
\begin{tabular}{
L{2.4cm}
C{0.6cm} C{0.7cm}
C{0.6cm} C{0.7cm}
C{0.6cm} C{0.7cm}
}
\toprule
& \multicolumn{2}{c}{Soccer}
& \multicolumn{2}{c}{Basket.}
& \multicolumn{2}{c}{Rock Cl.} \\
\cmidrule(lr){2-3} \cmidrule(lr){4-5} \cmidrule(lr){6-7}
\textbf{Method}
& $\bar{d}$ & Con.
& $\bar{d}$ & Con.
& $\bar{d}$ & Con. \\
\midrule
Random
& 0.0 & 50.0 & 0.0 & 50.0 & 0.0 & 50.0 \\
Video Cont.
& 0.03 & 51.2 & 0.13 & 53.0 & 0.15 & 53.5 \\
CrossTrainer \cite{crosstrainer}
& 0.04 & 54.7  & 0.16 & 54.3 & 0.19 & 60.4 \\
ZS VLM \cite{qwen25vl}
& 0.01 & 54.5  & 0.20 & 55.5 & 0.19 & 61.0 \\
Ours
& \textbf{0.20} & \textbf{64.4} & \textbf{0.25} & \textbf{56.3}  & \textbf{0.23} & \textbf{64.1} \\
\bottomrule
\end{tabular}
\end{minipage}
\label{tab:ranking_and_linearprobe}
\vspace{-0.5cm}
\end{table}

We first discuss the visual encoders used, and then show that using \OURS as training data improves the performance. We also show that the resulting representation gives a better linear probe performance. 

\noindent \textbf{Encoders.} We compare the performance of visual encoders trained with prior large-scale datasets with the same model trained with our dataset. We show this on two visual encoders and compare them for both a zero-shot and fine-tuned (with \OURS) setting.

\begin{itemize}
    \item \textbf{Image-based encoder}~\cite{clip}: We use CLIP-ViT-B-32~\cite{clip} as the starting model. The zero-shot baseline is the similarity between this visual encoder and the text representations~\cite{clip}. The clip representation is the average representations of the sampled frames from the video clip.
    \item \textbf{Video-based encoder}~\cite{internvideo2,timesformer}: We use InternVideo2.5~\cite{internvideo2} as the zero-shot video encoder. For training, we use a much lighter TimeSformer~\cite{timesformer} visual encoder. Note that since TimeSformer is a video-only backbone, it cannot be evaluated in a zero-shot setting.  
\end{itemize}

\begin{figure}[t]\footnotesize
    \centering
    
    \includegraphics[width=\linewidth]{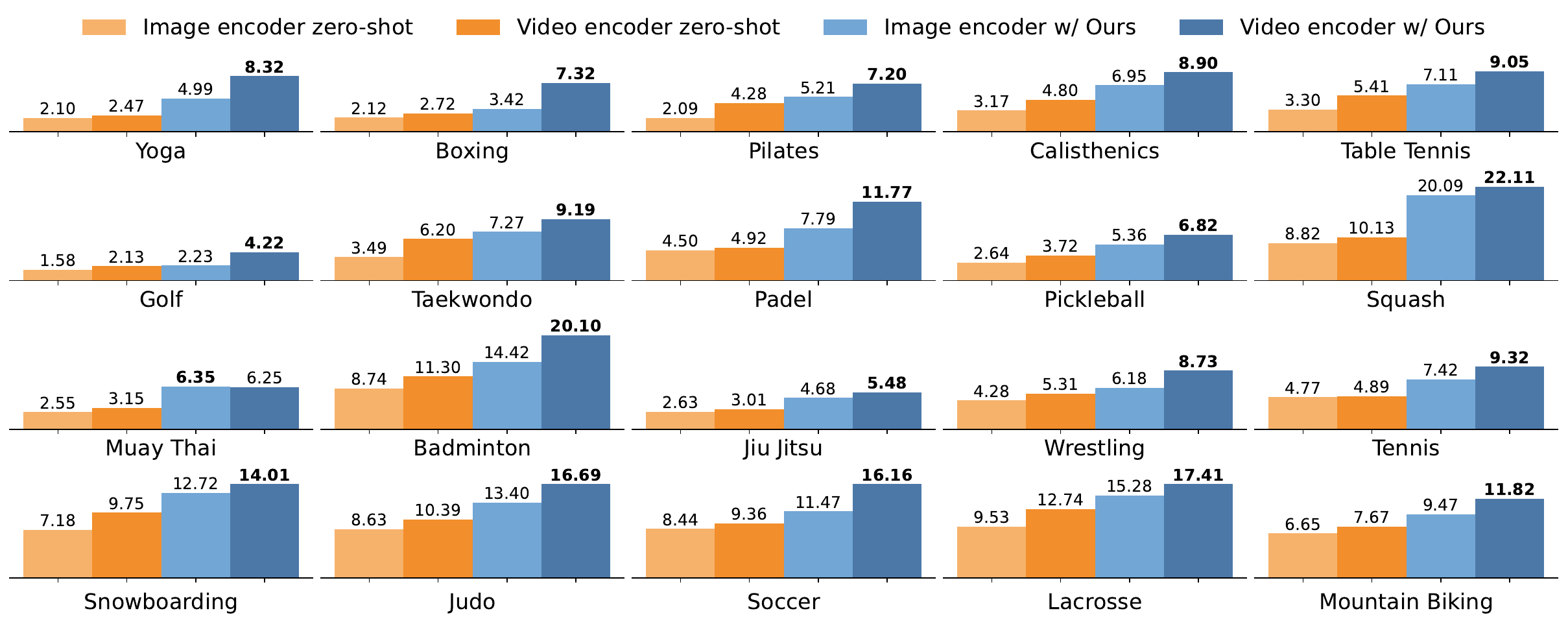}
    
    \caption{\textbf{Average recall$@10$ for 20 sports.} We show the performance of image and video-based visual encoder w/ and w/o \OURS as the training dataset (orange and blue shades, respectively). We see a clear increase in the encoders trained with our proposed dataset. See Appendix for the performance on the remaining sports. 
    }
    \label{fig:quantitative}
    \vspace{-0.5cm}
\end{figure}

\noindent \textbf{Evaluation metric.} For pretraining evaluation, we report retrieval metrics. We evaluate if the ground-truth $t$ is in the top-k retrieved narrations for a given video $v$, and vice-versa. We report mean recall$@10$, that is, the average of text-to-video recall$@10$, and video-to-text recall$@10$.

\noindent \textbf{Linear probe classification of correct vs incorrect demo.} To further showcase the usefulness of the learned representations, we evaluate the performance of these video representations on a simple linear probe classification task. Given a video $v$, we classify whether $v$ shows the demo positively, \ie $c=1$, or negatively, \ie $c=-1$. The idea is that the video representations should encode the `correct' ways of doing a skill, and the linear probe should reflect that.

\noindent \textbf{Results.} ~\cref{fig:quantitative} shows the results for $20$ sports (remaining in Appendix). We see a clear increase in mean recall$@10$ in the visual models trained with \OURS. The relative gains are as high as $\mathbf{4\times}$ compared to zero-shot models, \eg, Yoga. Across all sports, the performances of CLIP-ViT-B32~\cite{clip} (image encoder zero-shot), InternVideo2.5~\cite{internvideo2} (video encoder zero-shot), CLIP-ViT-B32 w/ \OURS (image encoder w/ Ours), and TimeSformer w/ \OURS (video encoder w/ Ours) are $11.05$, $11.34$, $12.75$, and $\mathbf{15.48}$, respectively. Moreover, training with \OURS improves the performance on $52/55$ sports. Overall, these results show the usefulness of physical skill-aware representations.

\begin{figure}[t!]\footnotesize
    \centering
    \includegraphics[width=\linewidth]{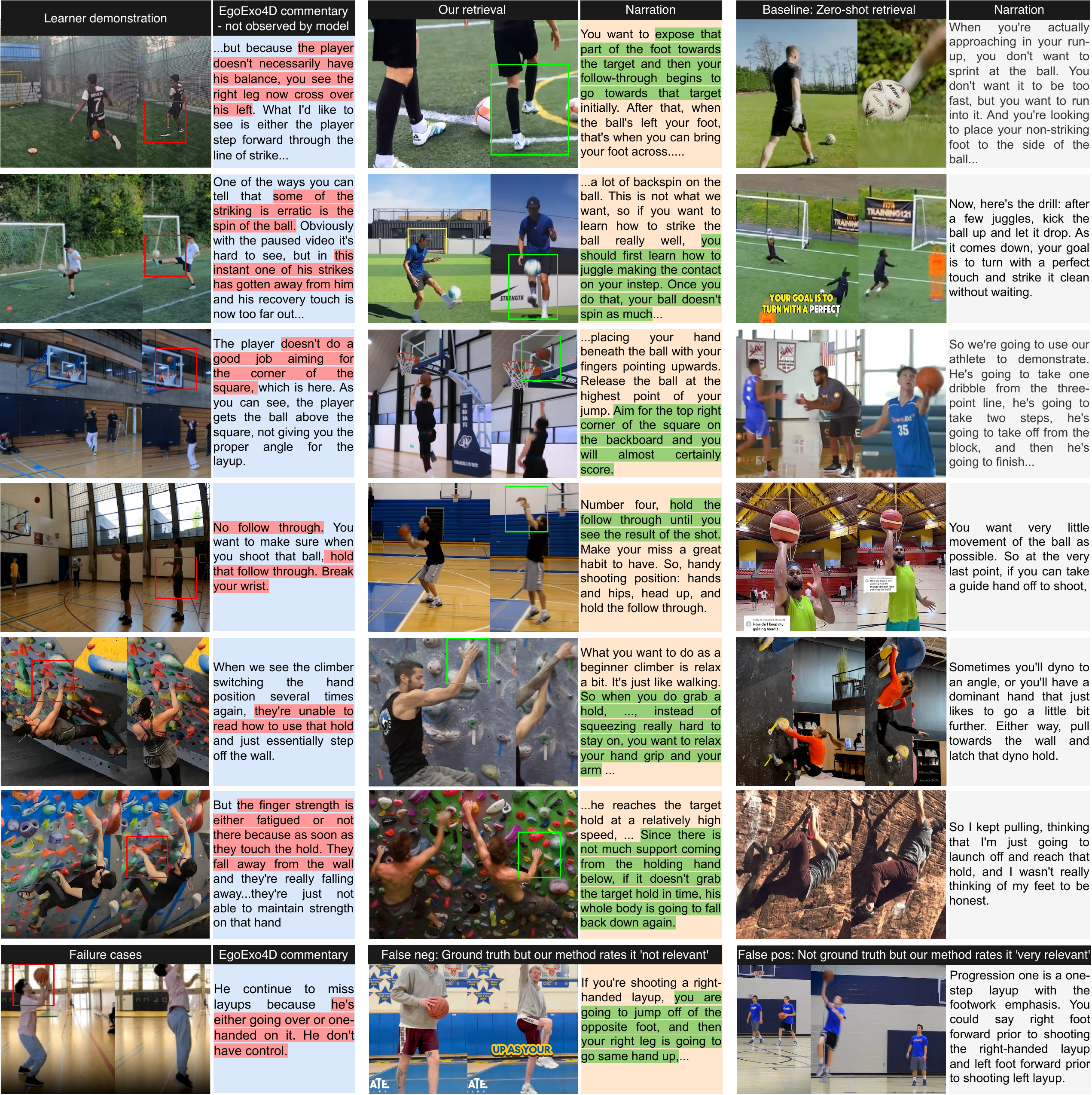}
    
    \caption{\textbf{Retrieval-based skill corrections.} We show representative outputs from our proposed model that provides visual feedback on a suboptimal learner performance, given only their video (left column). We do not input the expert commentary (second column) during either training or testing; it is given here for illustration. We see that the retrievals by our method provide personalized corrective feedback. On the other hand, retrievals from the zero-shot retrieval baseline are not helpful in skill improvement. (Last row) We also include a few failure cases that showcase the difficulty of the task. Recall that the learner videos are sourced from \cite{egoexo4d}, and the feedback videos are from our proposed \OURS. 
    }
    \label{fig:qualitative}
    \vspace{-0.5cm}
\end{figure}

\cref{tab:ranking_and_linearprobe} (left) shows the performance in linear probe classification for basketball, soccer, and rock climbing (see Appendix for others).  
We see significant gains in each of them, with gains up to $\mathbf{5\%}$. Linear probe on video features trained with \OURS outperforms linear probe on zero-shot video features. This experiment motivates the applicability of \OURS on various tasks requiring physical fine-grained skill understanding.

\subsection{Learning to provide visual feedback}
\label{sec:visual-feedback-training}

For this experiment, we first describe the baselines, following by the evaluation metric, and finally, the results.

\noindent \textbf{Baselines.} We compare our method against the following baselines:

\begin{itemize}
    \item Random: We randomly assign a confidence score for every $(d, v)$ pair.
    \item Zero-shot VLM~\cite{qwen25vl} + Zero-shot retrieval~\cite{mpnet}: In this baseline, we decompose obtaining $v$ from $d$ into two sub-problems: we generate actionable commentary $\hat{a}$ using VLM~\cite{qwen25vl}, \ie, generating $\hat{a}$ given $d$. Next, we take $\hat{a}$, and use sentence similarity~\cite{mpnet} to find the $t$ that is semantically similar to the predicted commentary $\hat{a}$. 
    \item CrossTrainer~\cite{crosstrainer} + Zero-shot retrieval~\cite{mpnet}: This baseline is a trained version of the previous zero-shot baseline. CrossTrainer~\cite{crosstrainer} is trained on EgoExo4D~\cite{egoexo4d} to generate better actionable feedback. Similar to the previous method, we obtain $\hat{a}$ using \cite{crosstrainer}, and use sentence similarity~\cite{mpnet} to obtain a semantically similar $d$. This method outperforms other trained methods to find $\hat{a}$~\cite{expertaf,qevd}.
    \item Video contrastive learning: This is a trained baseline where we contrastively match the learner demonstration $d$ directly to the instructional feedback narration $t$.
\end{itemize}

\noindent \textbf{Evaluation metrics.} In~\cref{sec:train-test-for-visual-feedback}, we introduce the gold-standard expert-labeled dataset CoachGT. 
To capture the correctness of $S$, we use two metrics. Firstly,  concordance rate~\cite{concordance} captures the fraction of time $S$ follows the same order of an expert's preference. In particular, for a given $d$, if an expert rated $v_1$ as being more relevant than $v_2$, then $S(d, v_1) > S(d, v_2)$ for a concordant sample. Next, we report average Cohen's $d$~\cite{cohen-d-1} that captures the difference in the score distributions of the three annotation classes in CoachGT.
Between the two sets of neighboring classes, we compute Cohen's $d$ as $\frac{\mu_1 - \mu_2}{\sigma}$ where $\mu_1$ and $\mu_2$ are the means of the adjacent classes, and $\sigma$ is the pooled standard deviation, and report the average value.
For this metric, higher values are better with $0$ as the random performance. 

\noindent \textbf{Results.}~\cref{tab:ranking_and_linearprobe} (right) shows the performance. Our method significantly outperforms all baselines on all sports. In particular, our method sees a gain of $\mathbf{10\%}$ compared to the strongest baseline in soccer, with a relative gain of $\mathbf{18\%}$. The higher concordance rate signifies correct relative ordering of our $S$, and the higher Cohen's $d$ value shows a good separation between the samples from distinct classes. Moreover, we observe that a simple contrastive training is not effective for this task. Nevertheless, even zero-shot feedback generation or CrossTrainer~\cite{crosstrainer} with semantic matching falls significantly short of our method's performance.

\cref{fig:qualitative} shows qualitative outputs from all sports (where Ego-Exo4D and \OURS overlap)---soccer, basketball, and rock climbing.  We see that our method retrieves instructional videos that correct the suboptimal executions by the learner. For example, the learner's soccer ball is moving erratically and spinning when attempting to juggle. In the suggested video, the coach asks to make the contact on the instep to avoid ball spin. This feedback is actionable and specific to the learner's issues. %
Recall that the sole input to the retrieval task is the learner video, no text.  These examples are representative of the behavior we generally see from the model.

We also show two failure cases in~\cref{fig:qualitative} (last row). Since our weakly-supervised dataset curation strategy uses sentence similarity, it is sensitive to the choice of words. In correcting the control when doing a layup, the suggested video explains better control without using the word `control', but our method gives a low score to this suggestion. Nevertheless, our proposed method enables leveraging \OURS as a source for visual feedback.

\section{Conclusion}
\label{sec:conclusion}

We propose \OURS, the first internet-scale dataset for fine-grained physical skill understanding. It contains more than $360k$ instructional videos from YouTube~\cite{youtube}, filtered to clips with instructional narrations that either visually demonstrate the correct execution of a fine-grained skill, or visually demonstrate what not to do. The dataset will serve as a foundation for future skill-learning research.
We demonstrate the value of learning from \OURS by showcasing an improved skill-aware fine-grained video representation %
as well as a novel visual-instructional feedback approach that converts a learner's video into personalized how-to demonstrations.   This paves the way for advances in
AI-based coaching and automated tutors for the physical world.

\bibliographystyle{splncs04}
\bibliography{main}

\appendix
\newpage

\title{\emph{Appendix for} \OURS: Physical Skill Learning from Sports Instructional Videos}
\author{Kumar Ashutosh, Chi Hsuan Wu, Kristen Grauman}
\authorrunning{Ashutosh et al.}
\institute{University of Texas at Austin}

\maketitle

\section{Overview of the Appendix}

This Appendix contains the following items:

\begin{itemize}
    \item A video summarizing our contributions, including samples from \OURS dataset, and results visualization.
    \item All prompts used with LLM~\cite{gpt4o} or VLM~\cite{qwen25vl}.
    \item Details of the annotation process for CoachGT, including the interface and examples.
    \item Additional experimental results.
\end{itemize}

\section{Language Model Prompts}
\label{sec:supp-prompts}

\textbf{Prompt for obtaining the list of sports.} We use the following prompt to generate a list of sports. We require the sports to be popular, involve physical skills, and be primarily performed by a single individual.

\begin{tcolorbox}[breakable, boxrule=0.2mm]
You are an expert in sports coaching and physical skill training.

Generate a list of 50 sports that satisfy the following requirements:

1. The sport should be reasonably popular and have a substantial amount of instructional or coaching content available online (e.g., tutorials, drills, or coaching videos).
2. The sport should involve physical skills where performance can be improved through visual, actionable feedback on body motion, technique, or form.
3. The sport should contain skills that are primarily performed by a single individual (e.g., dribbling in soccer, shooting in basketball), rather than skills that rely heavily on multi-player team coordination or strategy.

Avoid activities that primarily involve non-physical or non-visual concepts such as mentality, board games, or strategy-based games (e.g., chess).

Return only the list of sports as a numbered list. Each item should be the name of a sport.
\end{tcolorbox}

\noindent \textbf{Prompt for obtaining the skills set for each sport.} Next, we use the prompt given below to obtain the skills set for each sport. Since each sport has a unique set of skills, we extract the skills for each sport individually.

\begin{figure}[t]
    \centering
    \includegraphics[width=\linewidth]{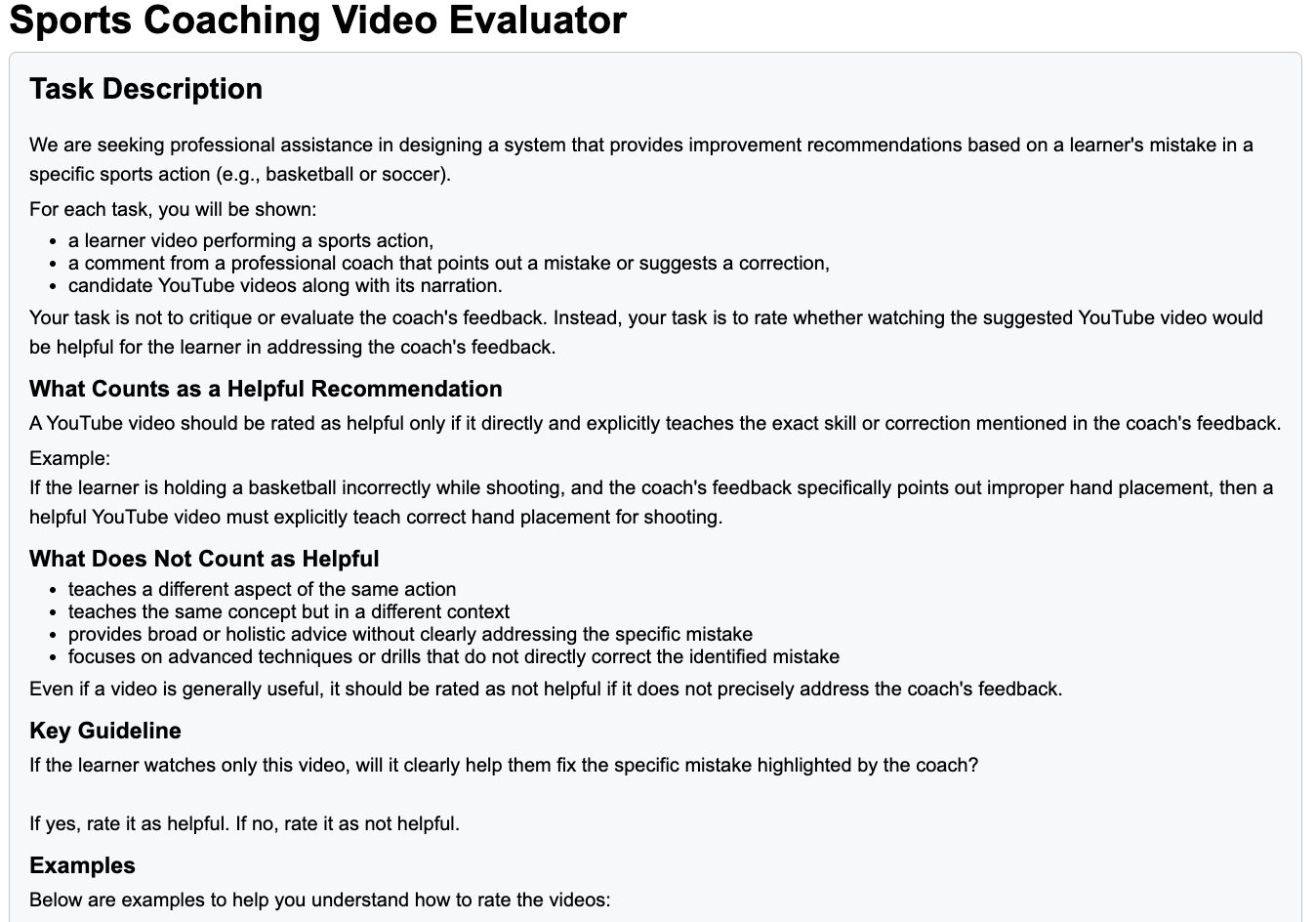}
    \caption{\textbf{Screenshot of the annotation instructions for CoachGT.} We provide a detailed description of the task to clarify the exact task to the experts.}
    \label{fig:annotation-interface}
\end{figure}

\begin{tcolorbox}[breakable, boxrule=0.2mm]

SYSTEM

You are an expert sports coach. Given the name of a sport, return a comprehensive list of all the fundamental and advanced skills required to master the sport. Include skills for all roles and positions (e.g., offense, defense, special roles). Mention only skills that are physical, and avoid non-visual skills like mentality. The list must be exhaustive but concise, avoiding duplicates. Output STRICT JSON ONLY with this schema:

\{ ``skills'': [``skill1'', ``skill2'', ...] \}

USER

Sport: \{sport\}
List all required skills exhaustively.

\end{tcolorbox}

\begin{figure}[t]
    \centering
    \includegraphics[width=\linewidth]{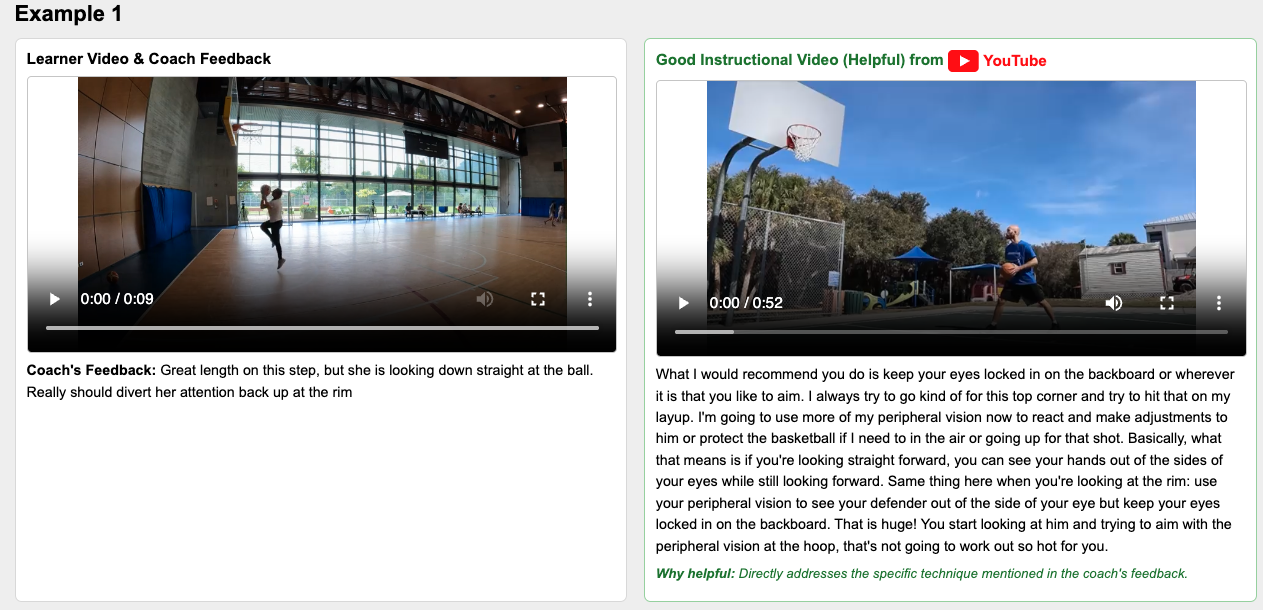}
    \includegraphics[width=\linewidth]{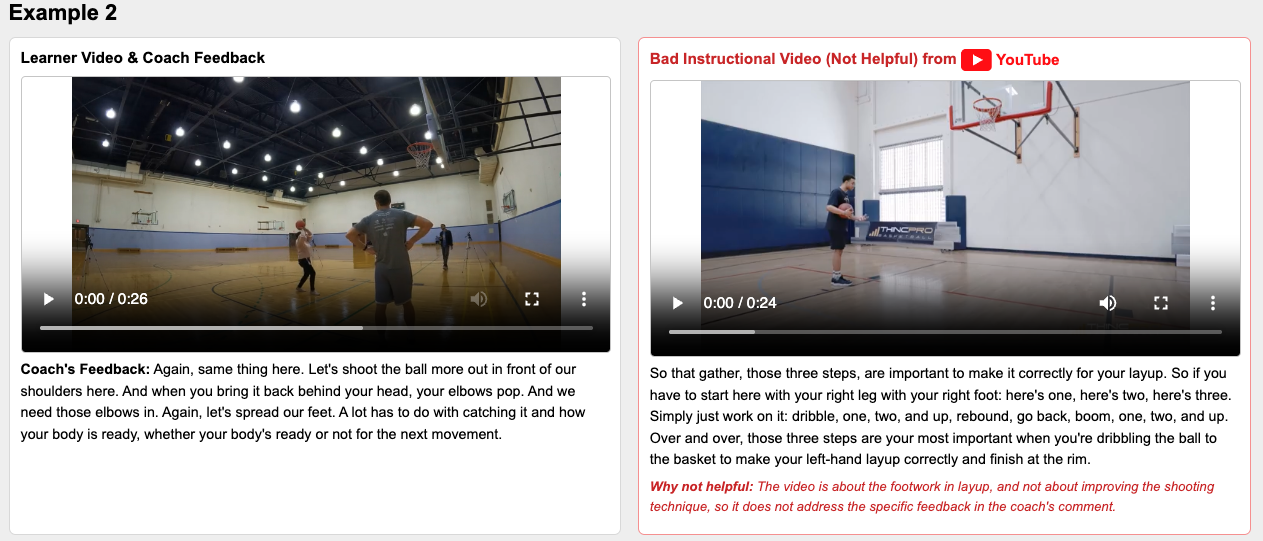}
    \caption{\textbf{Examples of the annotation questions for CoachGT.} We provide examples of the annotation questions for the experts to understand the task better.}
    \label{fig:annotation-examples}
\end{figure}

\noindent \textbf{LLM prompt for classifying if a commentary is actionable or not.} After we obtain narrations as text for every video, we filter out non-actionable instances, \eg, ``click the link to get discounts...'', \ie, narrations that surely do not contain visual actionable feedback on a skill. We use the following prompt:

\begin{tcolorbox}[breakable, boxrule=0.2mm]

SYSTEM

You classify whether a short narration contains actionable feedback on a skill.

ACTIONABLE: concrete, specific, and insight-based guidance on how to perform, fix, or improve a skill; contains steps, techniques, corrections, or criteria (not just motivation).

NOT-ACTIONABLE: motivational talk, intentions about future advice, vague statements, or mere descriptions.

Output STRICT JSON ONLY as:
\{ ``label'': ``actionable'' \} or \{ ``label'': ``not-actionable'' \}.

USER

\{narration comes here\}

\end{tcolorbox}

\begin{figure}[t]
    \centering
    \includegraphics[width=\linewidth]{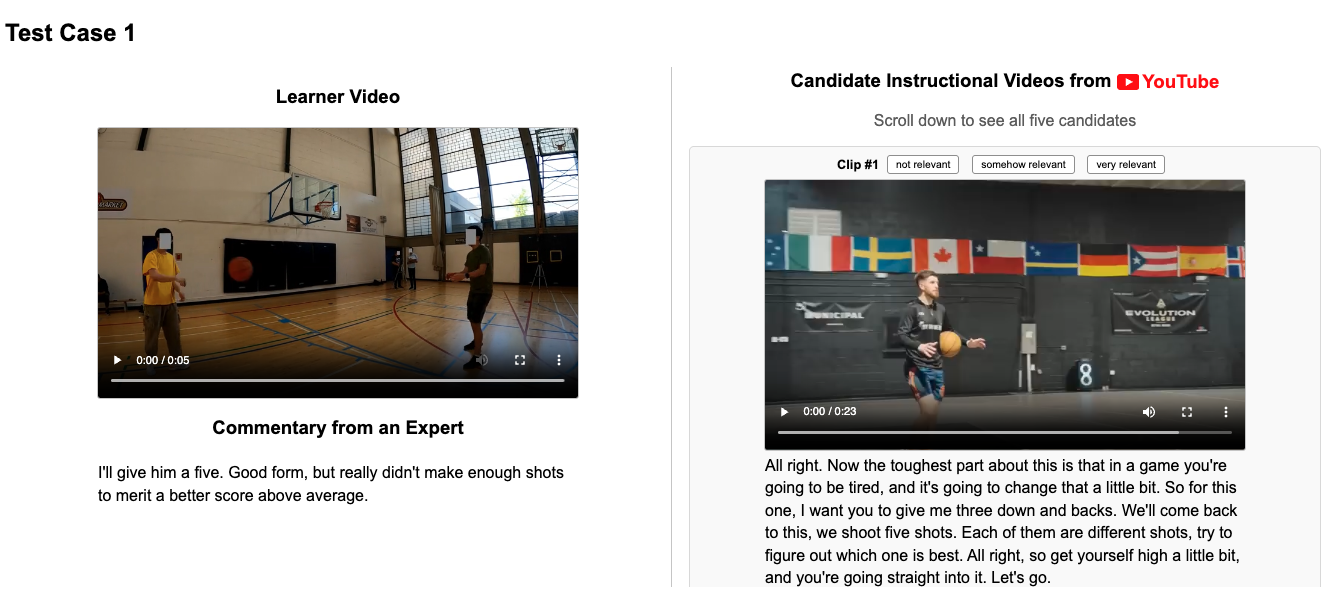}
    \caption{\textbf{Screenshot of the question.} We ask the expert to watch the learner video, read the expert feedback, and watch each of the five candidate corrective instructional videos to rate each of them as very relevant, somewhat relevant, or not relevant.}
    \label{fig:annotation-question}
\end{figure}

\noindent \textbf{VLM prompt to classify a visual demonstration and narration.} For the last stage of data preparation, 
we input the video clip and narration text and use the following prompt
to classify if the action demonstrated in the video shows how to do an action correctly, incorrectly, or not at all.

\begin{tcolorbox}[breakable, boxrule=0.2mm]

You are given a short coaching video clip and its narration text. 
Decide whether the action demonstrated in the video matches the narration (``Correct''), 
shows the wrong way of doing it (``Incorrect''), or if there is no skill demonstration at all (``None'').

Rules:

\begin{itemize}
    \item ``Correct'' → the action shown follows the narration or is explicitly marked as correct.
    \item ``Incorrect'' → the action shown demonstrates the wrong way of doing it, is explicitly marked as wrong (e.g., a cross $\times$ symbol on screen), or contradicts the narration (e.g., narration says ``do not spread legs'' 
  but the person spreads their legs).
  \item ``None'' → no skill demonstration is shown (e.g., only coach talking, headshots, unrelated visuals, or real 
  match footage with many players where the demonstration is unclear).
\end{itemize}

Additional Clarifications:

\begin{itemize}
    \item Do not classify as Correct or Incorrect if the clip only shows talking, close-up headshots, or unrelated visuals.
    \item Do not classify as Correct or Incorrect if the scene is a real match/game clip with many people and no clear step demonstration.
    \item If explicit visual cues like a cross $\times$ or text indicate "wrong way," classify as Incorrect even if the 
  action looks similar to the narration.
\end{itemize}

Output only one of the following words: Correct, Incorrect, or None. 
Add a sentence after a newline explaining your decision.

Here is the narration and the video: ``\{narration and video come here\}''.
    
\end{tcolorbox}

\begin{figure}[t]
    \centering
    \includegraphics[width=\linewidth]{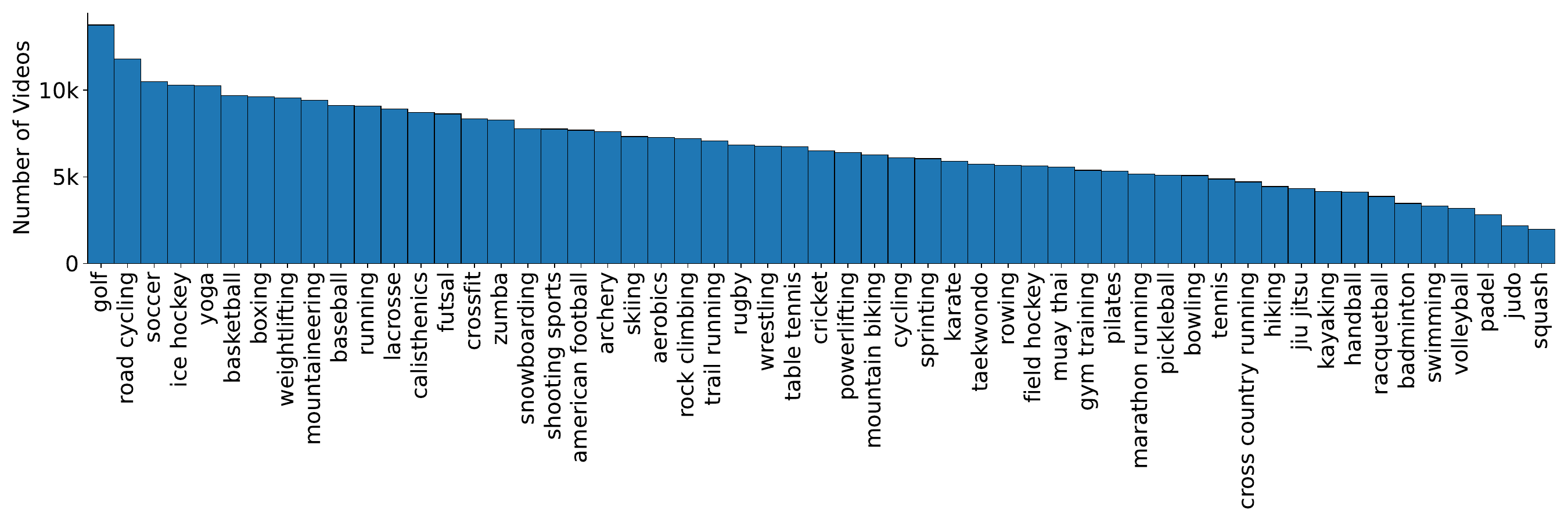}
    \caption{\textbf{Histogram plot} of the number of videos in \OURS per sport. We see that the number of videos per sport ranges from $2k$ to $14k$ with $6.7k$ videos on average.}
    \label{fig:histogram}
\end{figure}

\noindent \textbf{Prompt for training S'.} Lastly, we use the following prompt to train the model $S'$ for matching a learner video with an instructional narration. $S'$ is fine-tuned in LoRA setting.

\begin{tcolorbox}[breakable,boxrule=0.2mm]

You are an expert sports coach evaluating whether a video clip and an instructional text describe the same technique or skill.

Watch this video clip of someone practicing {sport}:
<video>

Now consider this instructional text:
``<narration>''

Based on the video and the instructional text, rate how well they match on a scale from 0 to 1:
\begin{itemize}
    \item 0: Completely unrelated (different skills, different techniques)
    \item 0.5: Somewhat related (same sport but different specific techniques) 
    \item 1: Strong match (same technique, similar body movements, related coaching points)
\end{itemize}

Output ONLY a single number between 0 and 1 representing the match score.
    
\end{tcolorbox}

\section{CoachGT Construction Details}

We hire six experts for the construction of CoachGT---two each for soccer, rock climbing, and basketball. The experts are professional coaches with more than $10$ years of experience. On average, the experts spend $4$ minutes on each annotation task, \ie, $15$ samples per hour. Since we ask them to rate the relevance of $5$ instructional clips per learner clip, we obtain a total of $75$ samples per hour per expert. In total, we annotate $~6$ hours of video clips, resulting in $450$ samples for rock climbing and basketball, and $451$ in soccer.

\begin{figure}[t!]
    \centering
    \includegraphics[width=\linewidth]{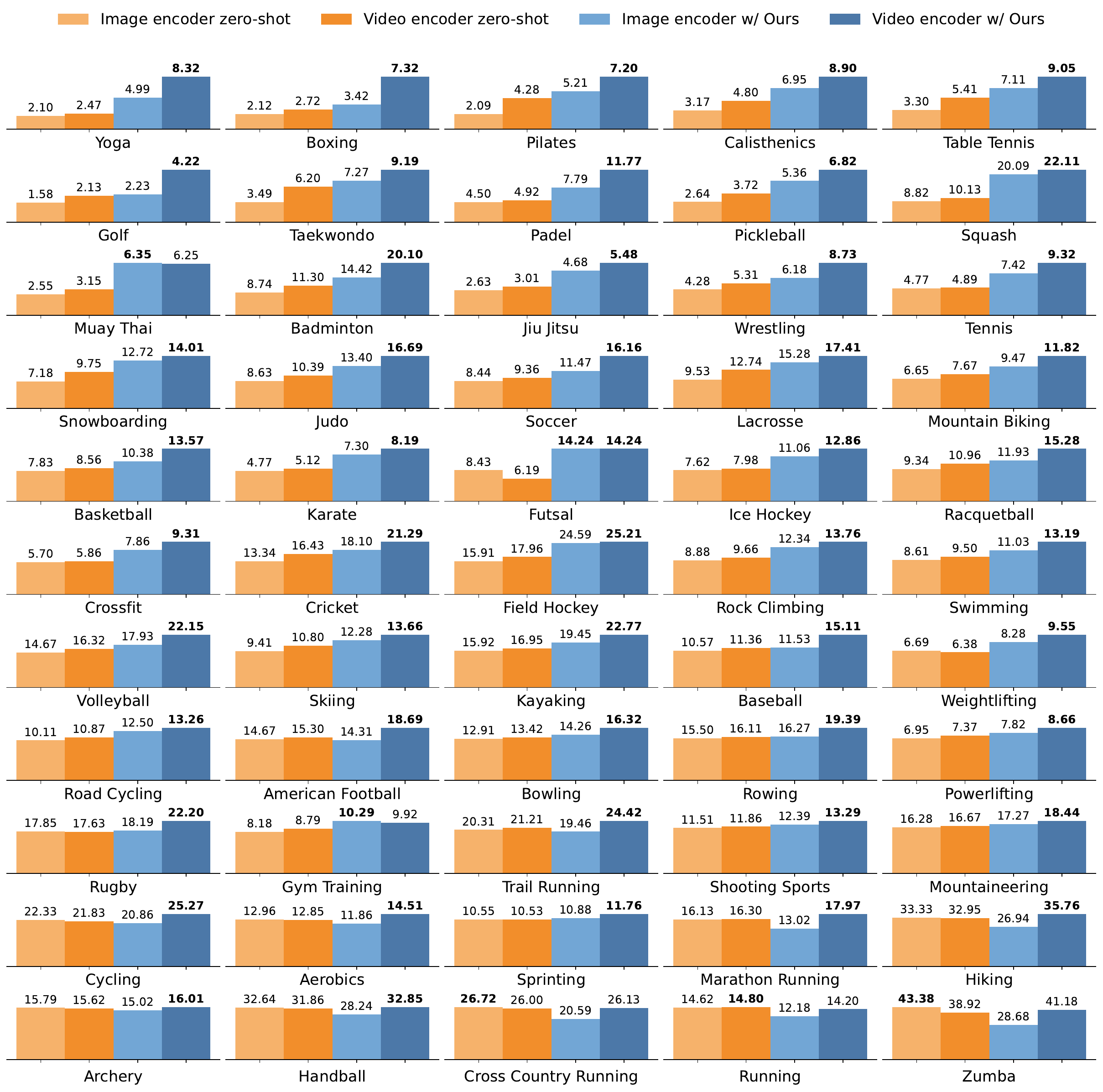}
    \caption{\textbf{Average recall$@10$ for all sports.} We show the performance of image and video-based visual encoders w/ and w/o \OURS as the training dataset (orange and blue shades, respectively). The light colors (blue and orange) represent image encodder performances, whereas the dark colors show the performance in the video encoders. We see a clear increase in the encoders trained with our proposed dataset. Note that even our image-encoder variant largely outperforms the video variant.}
    \label{fig:pretraining-performance-all-sports}
\end{figure}

The annotation interface is shown in \cref{fig:annotation-interface}. We show two examples for every sport, as shown in \cref{fig:annotation-examples}. The exact question to the annotator is shown in \cref{fig:annotation-question}. We will release the CoachGT test set to enable future research in this direction.

\section{More Details on \OURS and Results}

\subsection{Dataset distribution}

\cref{fig:histogram} shows the distribution of the number of videos for each sport in \OURS dataset. We see sports like golf and road cycling have more than $14k$ videos, whereas sports like judo and squash are around $2k$ videos. On average, we have $6.7k$ videos per sport.

\subsection{Physical skill-aware representation learning}

\cref{fig:pretraining-performance-all-sports} shows the mean recall@10 performance beyond the $20$ sports given in the main paper. We observe that training with \OURS as the training data improves the performance on $52/55$ sports. Specifically, \emph{zumba} has only $429$ video clips, and hence, we do not see any training improvement. Overall, our strong pretraining performance showcases the ability of video models to learn skill-aware representations with \OURS as the training data.

\subsection{Linear probe classification}

\cref{tab:supp_table} shows the usefulness of the learned video representations. It is an extended version of Tab.~\ref{tab:ranking_and_linearprobe} in the main paper. Due to compute constraints, we do this experiment on a random subset of $9$ sports, out of $55$.

We see that linear probe training on our physical skill-aware features gives better performance than the baseline for all sports. This result emphasizes the usefulness of skill-aware video representations.

\begin{table}[t!]
\centering
\footnotesize
\setlength{\tabcolsep}{3pt}

\caption{\textbf{Linear probe PR-AUC} of classifying a demonstration as correct or incorrect 
for a random subset of $9$ out of $55$ sports. We compare CLIP ViT-B/32 with the same model trained with the \OURS dataset. Our learned representation gives better results for all the chosen sports.
}

\begin{minipage}[t]{0.33\linewidth}
\centering
\begin{tabular}{
L{1.4cm}
C{0.9cm} C{0.9cm}
}
\toprule
\textbf{Sport} & ViT-B/32 & w/ Ours \\
\midrule
Soccer   & 58.2 & \textbf{63.2} \\
Basketball  & 60.2 & \textbf{64.9} \\
Rock Cl. & 56.5 & \textbf{61.2} \\
\bottomrule
\end{tabular}
\end{minipage}\hfill
\begin{minipage}[t]{0.33\linewidth}
\centering
\begin{tabular}{
L{1.4cm}
C{0.9cm} C{0.9cm}
}
\toprule
\textbf{Sport} & ViT-B/32 & w/ Ours \\
\midrule
Golf   & 61.2 & \textbf{65.7} \\
Futsal  & 58.0 & \textbf{61.2} \\
Hiking & 56.5 & \textbf{59.3} \\
\bottomrule
\end{tabular}
\end{minipage}\hfill
\begin{minipage}[t]{0.33\linewidth}
\centering
\begin{tabular}{
L{1.4cm}
C{0.9cm} C{0.9cm}
}
\toprule
\textbf{Sport} & ViT-B/32 & w/ Ours \\
\midrule
Badminton   & 57.2 & \textbf{60.1} \\
Yoga  & 54.7 & \textbf{59.9} \\
Tennis & 58.1 & \textbf{64.6} \\
\bottomrule
\end{tabular}
\end{minipage}\hfill
\label{tab:supp_table}
\vspace{-0.5cm}
\end{table}

\end{document}